\documentclass[conference]{IEEEtran}
\IEEEoverridecommandlockouts
\usepackage{amsmath,amssymb,amsfonts}
\usepackage{algorithmic}
\usepackage[noline,ruled,algo2e,linesnumbered]{algorithm2e}
\usepackage{comment}
\usepackage{multirow}
\usepackage{hyperref}
\usepackage{graphicx}
\usepackage{subcaption}
\usepackage{textcomp}
\usepackage{soul}
\usepackage{layouts}
\usepackage[table,xcdraw]{xcolor}

\newcommand{\cC}{\mathcal{C}} %
\newcommand{\cI}{\mathcal{I}} %
\newcommand{\cR}{\mathcal{R}} %
\newcommand{\cX}{\mathcal{X}} %

\def\BibTeX{{\rm B\kern-.05em{\sc i\kern-.025em b}\kern-.08em
    T\kern-.1667em\lower.7ex\hbox{E}\kern-.125emX}}

\usepackage{todonotes}

\newcommand{\ucb}{\text{UCB}} %
\newcommand{\qucb}{q\text{-UCB}} %

\makeatletter
\newcommand{\newlineauthors}{%
  \end{@IEEEauthorhalign}\hfill\mbox{}\par
  \mbox{}\hfill\begin{@IEEEauthorhalign}
}
\makeatother

\begin{document}

\bstctlcite{IEEEexample:BSTcontrol}

\title{Asynchronous Decentralized Bayesian Optimization for Large Scale Hyperparameter Optimization}


\author{\IEEEauthorblockN{Romain Egel\'e}
\IEEEauthorblockA{\textit{Universit\'e Paris-Saclay, France} \&\\
\textit{Argonne National Laboratory, USA}\\
romain.egele@universite-paris-saclay.fr}
 \and
\IEEEauthorblockN{Isabelle Guyon}
\IEEEauthorblockA{\textit{LISN, U. Paris-Saclay, France} \\
\textit{Google, USA}\\
guyon@chalearn.org}
 \newlineauthors
 \IEEEauthorblockN{Venkatram Vishwanath}
 \IEEEauthorblockA{\textit{Leadership Computing Facility} \\
 \textit{Argonne National Laboratory, USA}\\
 venkat@anl.gov}
\and
\IEEEauthorblockN{Prasanna Balaprakash}
\IEEEauthorblockA{\textit{Computing and Computational Sciences Directorate} \\
\textit{Oak Ridge National Laboratory, USA}\\
pbalapra@ornl.gov}
}


\maketitle
\thispagestyle{plain}
\pagestyle{plain}

\global\csname @topnum\endcsname 0
\global\csname @botnum\endcsname 0

\begin{abstract}
Bayesian optimization (BO) is a promising approach for hyperparameter optimization of deep neural networks (DNNs), where each model training can take minutes to hours. In BO, a computationally cheap surrogate model is employed to learn the relationship between parameter configurations and their performance such as accuracy. Parallel BO methods often adopt single manager/multiple workers strategies to evaluate multiple hyperparameter configurations simultaneously. Despite significant hyperparameter evaluation time, the overhead in such centralized schemes prevents these methods to scale on a large number of workers. We present an asynchronous-decentralized BO, wherein each worker runs a sequential BO and asynchronously communicates its results through shared storage. We scale our method without loss of computational efficiency with above 95\% of worker's utilization to 1,920 parallel workers (full production queue of the Polaris supercomputer) and demonstrate improvement in model accuracy as well as faster convergence on the CANDLE benchmark from the Exascale computing project.
\end{abstract}

\begin{IEEEkeywords}
machine learning, hyperparameter optimization, Bayesian optimization, asynchronous parallel computing
\end{IEEEkeywords}

\section{Introduction}

\begin{figure}[t]
    \centering
    \includegraphics[width=0.8\columnwidth]{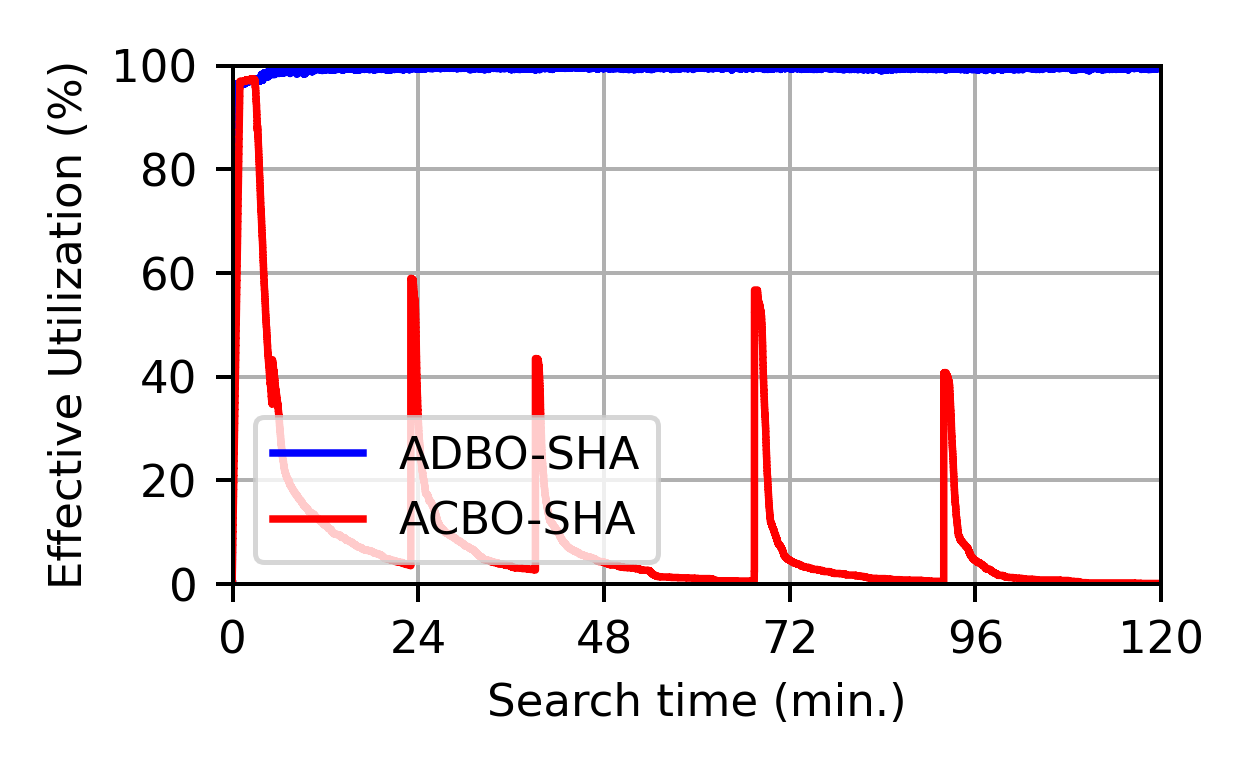}
    \vspace{-0.4cm}
    \caption{Utilization of computational resources between {\color[HTML]{CB0000}centralized (red)} and {\color[HTML]{3166FF}decentralized (blue)} Bayesian optimization equipped with Successive Halving (SHA) when using 1,920 workers (GPUs) (full Polaris HPC-system at the Argonne Leadership Computing Facility) to train neural networks configurations in parallel. {\color[HTML]{CB0000} The widely used single manager multiple worker utilization approach suffers from poor utilization}. {\color[HTML]{3166FF} Our proposed decentralized approach maintains high utilization}.}
    \label{fig:cbosha-vs-dbosha-utilization}
    \vspace{-0.4cm}
\end{figure}

Black-box optimization seeks to optimize a function based solely on input-output information. This problem is of particular interest in many scientific and engineering applications and is quite relevant to several machine-learning tasks. In the former case, the optimized black-box is often the result of a complex simulation code, software, or workflow where we can get only the output for a given input configuration. In the latter, many learning algorithms are sensitive to hyperparameters, which cannot be inferred during the training process and often need to be adapted by the user based on the training data~\cite{10.5555/3044805.3044891}. 
Existing methods for solving black-box optimization can be grouped into model-based and model-free methods. In the former, a surrogate model for the black-box function is learned in an online fashion and used to speed up the search~\cite{Wild_2015,hutter2011sequential,bergstra_algorithms_2011,hauschild2011introduction}. 
In the latter, the search navigates the search space directly without any explicit model~\cite{olsson1975nelder,poli2007particle,de2005tutorial,back1993overview,rutenbar1989simulated}. 
These two groups of methods have their own strengths and weaknesses. A key advantage of model-based over model-free methods is the sample efficiency w.r.t. the number of black box evaluations required by the search. Given the surrogate model, the search can quickly identify promising regions of the search space and find high-quality solutions faster (w.r.t. search iterations) than model-free methods can~\cite{shahriari_taking_2016}. 

Bayesian optimization (BO) is a promising class of sequential optimization methods. It has been used in a wide range of black-box function optimization tasks~\cite{shahriari_taking_2016,bischl2017mlrmbo,bartz2016survey}. In BO, an incrementally updated surrogate model is used to learn the relationship between the inputs and outputs during the search. The surrogate model is then used to prune the search space and identify promising regions.  BO navigates the search space by achieving a balance between exploration and exploitation to find high-performing configurations. While the exploration phase samples input configurations that can potentially improve the accuracy of the surrogate model, the exploitation phase samples input configurations that are predicted by the model to be high performing.

With the transition of high-performance  computing (HPC)  systems from petascale to exascale~\cite{10.1145/3372390}, massively parallel Bayesian optimizations that can take advantage of multiple computing units to perform simultaneous black-box evaluations are drawing attention to accelerate black-box optimization. These methods will be particularly beneficial for many HPC use cases, such as simulator calibration, software tuning, automated search of machine learning (ML) pipelines, neural network architecture and hyperparameter tuning (studied in this paper), and scientific simulation optimization. However, one of the main challenges is to transform commonly proposed sequential  BO~\cite{10.5555/2986459.2986743,hutter2011sequential,bergstra2013making,klein2017fast} algorithms to be parallel while keeping a similar sample efficiency.

The most sample-efficient heuristic for parallel BO methods is multipoint acquisition strategy under a centralized architecture, 
where the manager performs BO and workers evaluate the black-box functions. Also, these methods often use a Gaussian process regression (GPR) as a surrogate model~\cite{shahriari_taking_2016,frazier_tutorial_2018}. However, these two components respectively the centralized architecture and the GPR are two major bottlenecks for scaling BO in an HPC setting.

The problem we seek to solve is to {\bf improve both solution quality and the speed of hyperparameter optimization (HPO)}, by optimizing {\bf resource utilization of a large number of parallel computing workers}. To that end, we develop a parallel BO method based on a {\bf decentralized architecture} (without a single manager). Each worker runs its own sequential BO and {\bf communicates asynchronously} its hyperparameter evaluation results with other workers through a shared storage system. Additionally, each worker performs {\bf asynchronous hyperparameter suggestion for the next evaluation}.
Fig \ref{fig:cbosha-vs-dbosha-utilization} shows the comparison between the centralized and our proposed approach for neural network hyperparameter tuning from the ECP Candle benchmark. Despite each evaluation taking several minutes, the centralized scheme suffers from poor utilization at scale. Our proposed decentralized method overcomes the limitation of single manager scheme and achieves high resource  utilization. In Sec-\ref{sec:results}, we will show that the higher utilization results in superior solution quality as well.  

From the methodology perspective, our original contribution is to combine key algorithmic ingredients (decentralized, asynchronous surrogate model queries, $\qucb$ acquisition function, periodic exponential decay) which benefit from increasing the number of workers to improve the final solution as well as reduce the time to this solution.

From the software perspective, our original contribution is first to provide an abstract interface to submit and gather black-box evaluations in combination with a shared storage service. Our job scheduler can leverage different backends for distributed computing (e.g., threads, processes, MPI, Ray), similarly our storage service can leverage different data services (e.g., local memory, Redis, Ray Actors). We believe that this extensible architecture can open new research for HPC services tailored for large-scale hyperparameter optimization. Second, we provide a wrapper to launch seamlessly the method we propose with the Message Passing Interface (MPI) which could be valuable to run large-scale BO campaigns on many modern HPC systems.

Our {\bf principal findings} support the advantage of using asynchronous decentralized BO:
\begin{itemize}
    \item {\bf Stable resource utilization} when increasing the number of workers.
    \item {\bf Faster convergence and better solution} over the usual single manager and multiple workers when increasing the number of workers.
    \item {\bf Benefit from early discarding} strategies~\cite{8638041} to speed up the procedure at a fixed computational scale without loss of computational performance. 
    \item {\bf Parallel BO at HPC scale} involving 1,920 workers (one NVIDIA A100 GPU per worker).
\end{itemize}

\section{Related Work}
\label{sec:related-work}

Bayesian optimization is a well-established method to solve the global optimization problem of expensive and noisy black-box functions~\cite{mockus1978application}. For a detailed overview see~\cite{shahriari_taking_2016}. The problem we seek to solve is formally presented in Eq.~\ref{eq:genprob}:
\begin{equation}
\max_x \left\{ f(x) : x=(x_\cI,x_\cR,x_\cC) \in \cX \right\},
 \label{eq:genprob}
\end{equation}
where $x=(x_\cI,x_\cR,x_\cC)$ is a vector of $n_d$ parameters partitioned into three types of parameters $\cI$, $\cR$, and $\cC$, respectively denoting discrete parameters with a natural ordering, continuous parameters taking real values, and categorical parameters with no special ordering; $f(x)$ is a {\em computationally expensive} black-box objective function that needs to be maximized (minimization problems can also be carried out similarly by changing the sign of $f$). Typically, the feasible set $\cX$ is defined by a set of constraints on the parameters $x$. This includes bound constraints that specify the minimum and maximum values for the parameters and linear and non-linear constraints that express the feasibility of the given parameter configuration through algebraic equations. Consequently, given these algebraic constraints, the time required to verify the feasibility (that is if $x \in^?\cX$) is negligible. Hidden constraints are unknown to the user and generally require an evaluation of the black-box function $f$ to be uncovered. 
The objective function $f(x)$ can be deterministic (the same values $f(x)$ for the same $x$) or stochastic (different $f(x)$ values for the same $x$). Generally, finding the global optimal solution of the stated problem is computationally intractable~\cite{larson2019derivative,bartz2016survey,bischl2017mlrmbo}, except for the simplest cases. The presence of integer and categorical parameters, algebraic and hidden constraints, results in a discontinuous parameter space, which makes the search process difficult. Several mathematical optimization algorithms take advantage of gradients that measure the change in the value of the objective function w.r.t. the change in the values of the parameters. This is not feasible for the general problem \ref{eq:genprob}, however, because the black-box function cannot be differentiated. We focus on the optimization problem setting where the black-box function $f$ is computationally expensive to evaluate.

\subsection{Surrogate Model}
\label{sec:surrogate-model}

BO uses a surrogate model assumed to be computationally cheaper than the black-box in order to suggest the next points to evaluate. The choice of a good surrogate model plays a crucial role in the scalability and effectiveness of the BO search method. In most BO methods, GPR is employed because of its built-in uncertainty quantification capability~\cite{shahriari_taking_2016,frazier_tutorial_2018}. Specifically, GPR implicitly adopts Bayesian modeling principles of estimating the posterior distribution of output from the given input-output pairs and provides the predictive mean and variance for the unevaluated input configurations. GPR also has the advantage of being differentiable. However, while GPR is superior for faster convergence when run sequentially it remains one of the key bottlenecks for computational scalability in an HPC setting where thousands of input-output pairs (samples) can be computed in one batch. In fact, the GPR model needs to be refitted with a rapidly growing set of samples (past and new), but it has a cubic complexity $O(n_\text{sample}^3)$~\cite{liu_when_2019} w.r.t. the number of samples $n_\text{sample}$. For a small $n_\text{sample}$, this is not an issue. At scale, however, the cubic time complexity for model fitting will slow or even stop the search's ability to generate new input configurations, thus increasing the idle time of the workers and eventually resulting in poor HPC resource utilization as well as fewer overall  evaluations. Other surrogate models were proposed in the literature such as deep neural networks~\cite{snoek_scalable_2015}, Tree-Parzen estimation (TPE)~\cite{bergstra_algorithms_2011}, and random-forest regression (RFR)~\cite{breiman2001random,hutter_algorithm_2013}. We adopt RFR for its wide adoption thanks to its versatility with real, discrete, and categorical features as well as its robustness. RFR has a fitting time complexity of $O(n_\text{tree}\cdot n_\text{feature} \cdot n_\text{sample} \cdot \log(n_\text{sample}))$ with  $n_\text{tree}$ number of trees in the ensemble and $n_\text{feature}$ number of features ($=n_d$, problem dimension) per sample, which is constant for each search setting. In addition to the log-linear time complexity, RFR provides simple and easy in-node parallelization opportunities, where each tree can be built independently of other trees in the ensemble. It is also important to note that RFR is not sensitive to the re-scaling of the input space (features) but is particularly sensitive to the re-scaling of the target space. We apply a $\log(.)$ transformation on the normalized (between $[\epsilon, 1]$ up to a sign depending on maximization/minimization) target space to improve the convergence of the BO to quantities of interest.

\subsection{Multipoint Acquisition}
\label{sec:acq-func}

The way in which the input point $x$ is selected for evaluation is another bottleneck for scaling the number of workers. The selection method comprises an acquisition function that measures how good a point is and a selector that seeks to optimize the acquisition function over $\cX$. Typically, when all the parameters are continuous (or if they afford such encoding/transformation), specialized gradient-based optimizers are employed to select the next point. However, due to RFR not being directly differentiable, the mixed-integer nature of the parameter space, the presence of algebraic constraints, and the resulting discontinuity in the search space, such optimizers cannot be employed. While  specialized mixed-integer nonlinear solvers for these types of problems do exist, they are computationally expensive and cannot be employed in the fast iterative context required for scaling. Therefore, we use a point selection scheme with the upper confidence bound (\ucb)~\cite{shahriari_taking_2016} acquisition function on top of a random sample from $\cX$. This scheme selects an input point $x$ for evaluation as follows.  A large number of unevaluated configurations are sampled from the feasible set $\cX$. The BO uses a dynamically updated surrogate model $m$ to predict a point estimate (mean value) $\mu(x)$ and variance $\sigma(x)^2$ for each sampled configuration $x$. The sampled configurations are then ranked using the \ucb\  given by
\begin{equation} \label{eqn:ucb}
    \ucb(x) = \mu(x) + \kappa \cdot \sigma(x),
\end{equation}
where $\kappa \geq 0$ is a parameter that controls the trade-off between exploration and exploitation. When $\kappa$ is set to zero, the search performs only exploitation (greedy); when $\kappa$ is set to a large value, the search performs stronger exploration. A balance between exploration and exploitation is achieved when $\kappa$ is set to an appropriate value, classically $\kappa = 1.96$, which translates into  a 95\% confidence interval around the mean estimate when computing \ucb.

In parallel BO, there are mainly two ways for querying $q$ new suggestion in parallel. On the one hand, the centralized method tries to resolve a multipoint optimization problem such as the $q-\text{EI}$ criteria~\cite{ginsbourger2010kriging,shahriari_taking_2016}. In this case, a manager runs the BO and generates configurations~\cite{gonzalez_batch_2015,snoek_practical_2012}, and the workers  evaluate the configurations and return the results back to the manager. The manager generates configurations in a batch synchronous or asynchronous way~\cite{pmlr-v97-alvi19a}. But, to generate these batches the multipoint optimization problem has to be solved and it becomes harder and more computationally expansive  when $q$ increases. Therefore, heuristics such as the constant-liar (CL) strategy~\cite{ginsbourger2010kriging,balaprakash_deephyper_2018} have been proposed to approximate this criterion. Still, the CL scheme has a linear temporal complexity w.r.t. the number of workers which impacts performance when increasing the number of workers. That is, for $q$ new suggestion the surrogate model needs to be updated $q$ times sequentially. In addition, when a configuration finishes (at any time) the last batch queried can still be processed which provokes congestion in the manager's queue.
In contrast to the centralized architectures, decentralized BO was recently introduced based on stochastic policies such as Thompson sampling and Boltzmann policy~\cite{hernandez-lobato_parallel_2017,garcia2019fully}. Stochastic policies allow bypassing communication in the decentralized architecture, which enable asynchronous iterations. However, the asynchronous case was not studied in these works.

\section{Method}
\label{sec:method}

\begin{figure*} 
    \centering
    \begin{subfigure}{\columnwidth}
        \centering
        \includegraphics[width=0.6\textwidth]{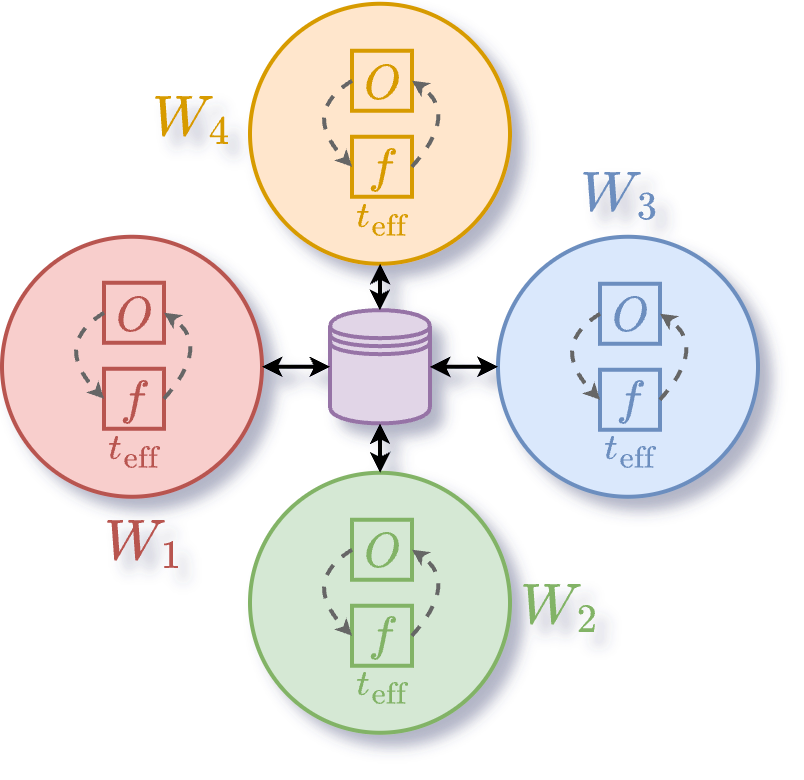}
        \caption{}
        \label{fig:distributed-search}
    \end{subfigure}
    \begin{subfigure}{\columnwidth}
        \centering
        \includegraphics[width=0.6\textwidth]{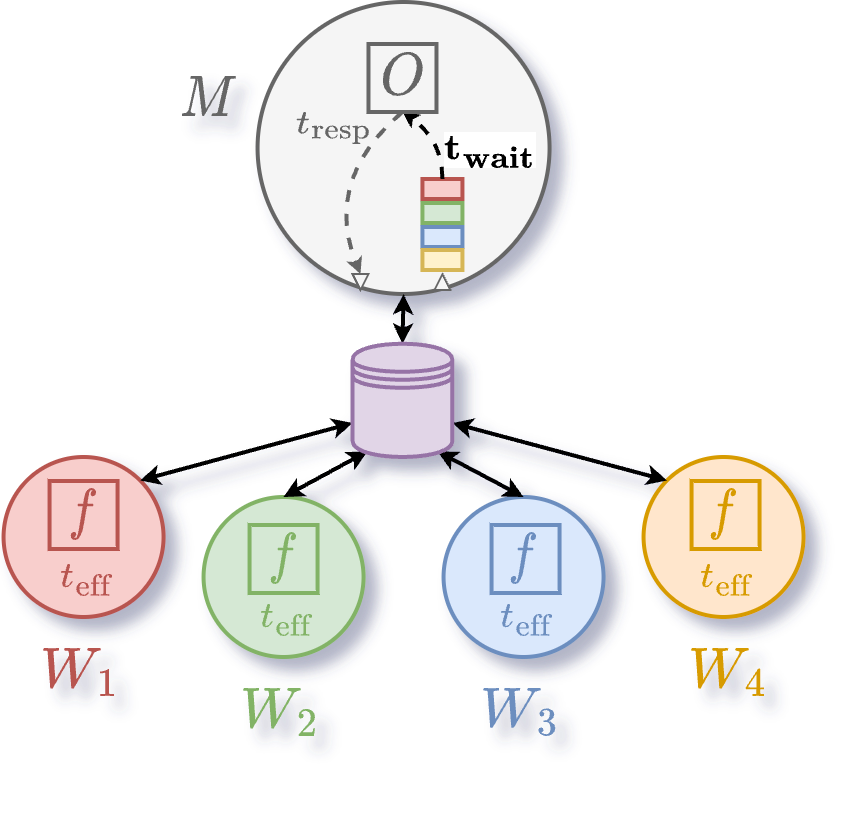}
        \caption{}
        \label{fig:centralized-search}
    \end{subfigure}
    \caption{Decentralized (\ref{fig:distributed-search}) and centralized (\ref{fig:centralized-search}) search models with a shared storage. A circle represents a process with $W$ for a worker and $M$ for a manager, an arrow represents a communication, $O$ represents the optimizer, and $f$ represents the computation of the black-box function. $t_{\text{wait}}$ is the time for which a worker waits before being processed by the optimizer. $t_\text{resp}$ is the time taken by the optimizer to suggest a new configuration.}
\end{figure*}

In this section, we introduce our novel approach to BO in the context of HPC exploiting a decentralized architecture with asynchronous communication. In fact, multiple compute units can be used on an HPC platform, where each function evaluation requires a fraction of this platform. Let $n_\text{worker}$ be the number of available workers, where a worker represents a unit of computational resource available to evaluate the black-box function (e.g., CPU, GPU, a fraction of a compute node, whole node, or a group of nodes). Let $T_{\text{wall}}$ be the total wall-clock time for which these resources are available (i.e., job duration). Then the overall available compute time $T_{\text{avail}} = n_\text{worker} \cdot T_{\text{wall}}$ upper bounds the total time spent in black-box function evaluations $T_{\text{eff}} = \sum_{t_\text{eff} \in \mathcal{T}} t_\text{eff}$ used to perform the job, where $\mathcal{T}$ is the set of duration $t_\text{eff}$ for all evaluated black-box functions. We define ``effective utilization'' as $U_{\text{eff}} = T_\text{eff} / T_\text{avail} $. For the problem of ``parallel black-box optimization'' we seek to maximize the objective function $f$, as well as maximize the effective utilization $U_\text{eff}$. We maximize utilization by minimizing the computational overhead of the search algorithm. In other words, in addition to the objective function maximization, the optimization method should effectively use parallel resources by maximizing their usage mainly with black-box evaluations.

BO is a promising approach for tackling the class of black-box optimization problems described in Eq.~\ref{eq:genprob}. BO tries to leverage accumulated knowledge of $f$ throughout the search by modeling it as a probability distribution $p(y|x)$, which represents the relationship between $x$, the input,  and $y$, the output. Typically, BO methods  rely on dynamically updating a surrogate model that estimates $p(y|x)$. Often, this distribution is assumed to follow a normal distribution. Therefore, the surrogate model estimates both $\mu(x)$ the mean estimate of $y$ and $\sigma(x)^2$ the variance. 
The latter is leveraged to assess how uncertain the surrogate model is in predicting $x$~\cite{shahriari_taking_2016}. The surrogate model is cheap for prediction and can be used to prune the search space and identify promising regions, where the surrogate model is then iteratively refined by selecting new inputs that are suggested by the model to be high-performing (exploitation) or that can potentially improve the quality of the surrogate model (exploration). BO navigates the search space by achieving a balance between exploration and exploitation to find high-performing input configurations. 

\subsection{Asynchronous decentralized Bayesian Optimization}
\label{sec:method-adbo}

Figure \ref{fig:distributed-search} shows a high-level sketch of our proposed asynchronous decentralized Bayesian optimization (ADBO) method. The key feature of our method is that each worker executes a sequential BO search; performs only one black-box evaluation, which avoids congestion occurring in a centralized setting (see Figure~\ref{fig:centralized-search}); and communicates the results with all other workers in an asynchronous manner through shared storage. The BO of each worker differs from that of the other workers w.r.t. the value $\kappa_0$ used in the $\ucb$ acquisition function. Each BO starts by sampling the value $\kappa_0$ from an exponential distribution $\text{exp}(\frac{1}{\kappa})$, where $\kappa$ is the user-defined parameter. The BO that takes smaller  $\kappa_0$ values will perform exploitation and sample points near the best found so far in observations. On the other hand, the BO that receives large $\kappa_0$ values will perform exploration to reduce the predictive uncertainty of the RFR model.
Consequently, on average multiple BO searches will seek to achieve a good trade-off between exploration and exploitation; however, there will be a number of BO searches that perform stronger exploitation or exploration. This effect will increase as we scale to a large number of workers and can benefit the overall search. Our approach is inspired by the $\qucb$ acquisition function~\cite{jones2001taxonomy,hutter2012parallel} where different $\kappa$ values are sampled from the exponential distribution for the $\ucb$ acquisition function and different points are selected based on these values to find the balance between exploration and exploitation. The main reason for adopting $\qucb$ is computational simplicity. Compared to other multipoint generation strategies, such as the constant liar~\cite{ginsbourger2010kriging} (denoted by CL) that exist only in the centralized (single-manager/multiple-worker) methods and has an overhead increasing linearly with the number of workers, there is only a constant overhead when using $\qucb$.

Algorithm \ref{alg:dmbs-process} shows the high-level pseudocode of our proposed ADBO method. The worker uses the \texttt{Storage}, a shared memory interface that keeps track of the input-output pairs seen by it and other workers. The search proceeds by initializing the optimizer object with a randomly sampled $\kappa_0$ value from an exponential distribution; the search is implemented through suggest-and-observe interfaces. The former implements the acquisition function and selector functionality and is used to generate an input $x_i$ for the evaluation of the black-box function. The latter is used to give the evaluation results and retrain the RFR surrogate model.  

As soon as the black-box evaluation is completed, the \texttt{write} function is used to share the $(x,y)$ with all other workers through the \texttt{Storage} (e.g. a database service). Next, the \texttt{read} function is used to read the new evaluations from other workers that were added to the \texttt{Storage} since it was last checked. Both \texttt{write} and \texttt{read} are asynchronous; the former does not wait for the acknowledgment of other workers, and the latter does not wait for all workers to send their most recent results. Local data and remote data are concatenated to update the surrogate model through the \texttt{observe} interface.

\begin{algorithm2e}[!t]
\small
\SetInd{0.25em}{0.5em}
\SetAlgoLined
\SetKwInOut{Input}{Inputs}\SetKwInOut{Output}{Output}
\SetKwFunction{Tell}{observe}
\SetKwFunction{Ask}{suggest}
\SetKwFunction{History}{History}
\SetKwFunction{Optimizer}{Optimizer}
\SetKwFunction{Storage}{Storage}
\SetKwFunction{Exp}{Exp}
\SetKwFunction{Update}{update}
\SetKwFunction{sendall}{send\_all}
\SetKwFunction{recvany}{recv\_any}
\SetKwFunction{bcast}{broadcast}
\SetKwFunction{recv}{receive}
\SetKwFunction{writef}{write}
\SetKwFunction{readf}{read}
\SetKwFunction{sample}{sample}

\SetKwFor{For}{for}{do}{end}

\Input{$f$: black-box function, $\kappa$: $\ucb$ hyperparameter, $rank$: rank of the process, $size$: number of processes, $seed$: random seed, $T$: period of the exponential decay, $\lambda$: decay rate of the exponential decay}
\Output{$\Storage$ all evaluated configurations}
    {\color{orange}\tcc{Initialization}}
    $\kappa_0 \leftarrow $ \Exp.\sample{$\frac{1}{\kappa}, size$}$[rank]$ \; 
    $optimizer \leftarrow $ \Optimizer{} \;
    $storage \leftarrow$ \Storage{} \;
    $t \leftarrow 0$\;
    {\color{orange}\tcc{Main loop}}
    \While{not done}{
        {\color{blue}\tcc{Apply exponential decay}}
        $\kappa_t \leftarrow \kappa_0 \cdot \exp(-\lambda \cdot (t \bmod T))$ \;
        ~\\
        $x \leftarrow$ $optimizer$.\Ask{$\kappa_t$} \;
        $y \leftarrow f(x)$ \;
        ~\\
        {\color{blue}\tcc{Write and read from the storage}}
        {$X, Y \leftarrow $\Storage.\readf{}} \;
        {$\Storage.\writef{x, y}$} \;
        $t \leftarrow t + 1$ \;
        ~\\
        {\color{blue}\tcc{Update the surrogate}}
        $optimizer.$\Tell{$X,Y$} \;
    }
    \Return \Storage\\
    \caption{Asynchronous Decentralized Bayesian Optimization (Worker Process)}
    \label{alg:dmbs-process}
\end{algorithm2e}

\subsection{Uncertainty Quantification in Random-Forest}
\label{sec:surrogate-model-uq}

Although RFR provides computational advantages, its uncertainty quantification capabilities are not well-known or documented in the literature. The most widely used RFR implementation is from the Scikit-Learn package~\cite{scikit-learn}. Our analysis of uncertainty quantification with the default implementation of this package showed that the predictive variance is not as good as that of GPR. The primary reason was the best-split strategy adopted in the usual random forest algorithm to minimize the variance of the estimator. Although this results in better predictive accuracy, the predictive variance is not informative in the context of Bayesian optimization because it remains constant in unexplored areas. We tested the random-split strategy for tree splitting, as suggested in~\cite{hutter_algorithm_2013}, and found that the uncertainty estimates from RFR are improved and are comparable to those with GPR in interpolation area while remaining constant in extrapolation areas. But because it does not follow the conventional RF algorithm, the split strategy is not exposed as a parameter to the user. We believe that this might be one of the reasons that RFR, despite its computational advantages, was not thoroughly experimented with for the purpose of uncertainty quantification. The uncertainty of the RFR is computed by applying the law of total variance on the mean estimate $\mu_\text{tree}(x)$ and variance estimate $\sigma_\text{tree}(x)^2$ of the trees of the ensemble, which are learned by minimizing the squared error:
\begin{equation} \label{eq:law-to-variance}
\sigma(x)^2 = \mathbb{E}[\sigma_\text{tree}(x)^2] + \mathbb{V}[\mu_\text{tree}(x)]
\end{equation}
where $\mathbb{E}[.]$ and $\mathbb{V}[.]$ are respectively the empirical mean and variance. We also compared the speed of our RFR with improved uncertainty (based on Scikit-Learn) against the Python package \texttt{pyrfr}\footnote{\url{https://github.com/automl/random_forest_run}} used by~\cite{hutter_algorithm_2013}. We observed that our implementation is orders of magnitude faster. For a test case with a single continuous input variable and a single continuous output target with 10,000 samples (2/3 training and 1/3 test), ours takes 0.12 seconds while the \texttt{pyrfr} implementation takes 24 seconds. The implementation based on Scikit-Learn can also benefit from multi-processing. Our new implementation of RFR with its improved uncertainty estimate module will be made available as  open-source software for future research.

\subsection{Periodic Exponential Decay}
\label{sec:periodic-exponential-decay}

We propose a heuristic to dynamically manage the trade-off between exploration and exploitation of the $\qucb$ in ADBO. For this, we rely on the idea of exponential decay for the $\kappa$ parameter of the $\qucb$ acquisition function, given by: 
\begin{equation}
    \kappa_{\lambda,T}(t;\kappa_0) = \kappa_0 \cdot \exp(-\lambda \cdot (t \bmod T)
\end{equation}
where $\kappa_0$ is the initial exploration-exploitation trade-off, $t$ is the number of local evaluations performed in the process, $T$ is the period of the scheduler and $\lambda$ is the decay rate of the scheduler. From this definition, in the case of ADBO, the $\kappa_0$ is a random variable sampled from an exponential distribution $\text{exp}(1/\kappa)$ where $\kappa$ is the mean of the distribution. Therefore we have the following average behavior for $\kappa$ across processes in the synchronized case:
\begin{equation}
    \mathbb{E}_{\kappa_0 \sim \text{Exp}(1/\kappa)}[\kappa_{\lambda,T}(t;\kappa_0)] = \kappa \cdot \exp(-\lambda \cdot (t \bmod T))
\end{equation}

With such a scheduler, the different processes are periodically converging to the ``exploitation'' regime (small $\kappa_t$ values) while the original ``exploration'' is regularly recovered. This heuristic avoids ``over-exploring'' (i.e., behavior similar to random search) when the number of parallel workers is increased.

\subsection{Asynchronous Successive Halving}
\label{sec:asynchronous-successive-halving}

We propose to combine our Bayesian Optimizer with an early rejection strategy. Early rejection strategies are methods used to discard early a hyperparameter configuration if it is not likely to improve over the best objective observed so far. In this work, we use the asynchronous successive halving (SHA) algorithm~\cite{li_system_2020}. The SHA algorithm compares the performance of a new configuration with past evaluations. This comparison is done at different "rounds" allocated according to a geometric progression (based on training epochs such as 1, 3, 9, 27...). An algorithm is stopped if it is not among the top-$\frac{1}{\rho} \times 100$\% at the current round where $\rho$ is the reduction factor.

\section{Results}
\label{sec:results}

We conduct an empirical study to demonstrate the benefits of using a decentralized architecture.
First, we analyze the effectiveness of the two approaches by looking at the number of idle resources as well as the number of completed neural network training for a fixed computational budget (i.e., number of workers and time of execution). Second, we analyze the gain in the final objective and the speed of the HPO.

{\bf Platform and main libraries --} All the experiments are performed on the Combo  benchmark from the Exascale-Computing Project. They are conducted on the Polaris supercomputer at the Argonne Leadership Computing Facility (ALCF). Polaris is a HPE Apollo Gen10+ platform that comprises 560 nodes, each  equipped with a 32-core AMD EPYC "Milan" processor, 4 Nvidia A100 GPUs, and 512 GB of DDR4 memory. The compute nodes are interconnected by a Slingshot network. In our experiments, each worker is attributed 1 GPU. The algorithm is implemented in Python 3.8.13 where the main packages used are deephyper 0.5.0, mpi4py 3.1.3, scikit-learn 1.1.2, Redis 7.0.5. Neural networks are implemented in Tensorflow 2.10. The number of workers is increased from 40 (10 nodes), 160 (40 nodes), 640 (160 nodes) to 1920 (480 nodes -- the full system in production queue).

{\bf Benchmark --} The Combo benchmark dataset~\cite{xia_predicting_2018} is composed of 165668 training data points (60\%), 55,222 (20\%) validation data points and 55,222 (20\%) test data points respectively. Each data point has three types of input features: 942 (RNA-Sequence), 3,839 (drug-1 descriptors), and 3,893 (drug-2 descriptors) respectively. The data set size is about 4.2 GB. It is a regression problem with the goal of predicting the growth percentage of cancer cells given a cell line molecular features and the descriptors of two drugs. The networks are trained for a maximum budget of 50 epochs and 30 minutes. The validation $R^2$ coefficient at the last trained epoch is used as the objective for hyperparameter search.

The baseline model is composed of 3 inputs each processed by a sub-network of three fully connected layers. Then, the outputs of these sub-models are concatenated and input in another sub-network of 3 layers before the final output. All the fully connected layers have 1000 neurons and ReLU activation. It reaches a validation and test $R^2$ of 0.87 after 100 epochs.
The number of neurons and the activation function of each layer are exposed for the hyperparameter search. The search space is defined as follows: the number of neurons in $[10, 1024]$ with a log-uniform prior; activation function in [elu, gelu, hard sigmoid, linear, relu, selu, sigmoid, softplus, softsign, swish, tanh]; optimizer in [sgd, rmsprop, adagrad, adadelta, adam]; global dropout-rate in $[0, 0.5]$; batch size in $[8, 512]$ with a log-uniform prior; and learning rate in $[10^{-5}, 10^{-2}]$ with a log-uniform prior. A learning rate warmup strategy is activated based on a boolean variable. Accordingly, the base learning rate of this warmup strategy is searched in $[10^{-5}, 10^{-2}]$ with a log-uniform prior. Residual connections are created based on a boolean variable. A learning rate scheduler is activated based on a Boolean variable. The reduction factor of this scheduler is searched in $[0.1, 1.0]$, and its patience in $[5, 20]$. An early-stopping strategy is activated based on a Boolean variable. The patience of this strategy is searched in $[5, 20]$. Then, batch normalization is also activated based on the Boolean variable. The loss is searched among [mse, mae, logcosh, mape, msle, huber]. The data preprocessing is searched among [std, minmax, maxabs]. This corresponds to 22 hyperparameters. All experiments are performed with the same initial random state 42.

{\bf Metrics --} During HPO the objective maximised is the coefficient of determination $R^2$. Then, for our analysis, we use the regret defined as $1-R^2$ where 1 is the upper bound of the objective. The model selection is always based on the validation scores but the test scores are presented to consider the problem of generalization in ML. To compare the speed of convergence between experiments as well as the quality of the solution we compute the Area Under the Regret Curve (AURC) defined in~\cite{liu_winning_2020} but without re-scaling of the time. The smaller the AURC the best is the any-time performance. Both the regret and the AURC are common metrics used in AutoML~\cite{liu_winning_2020,eggensperger2021hpobench}.

\subsection{Resource utilization}
\label{sec:exp-computational-performance}

\begin{figure}[!t]
    \centering
    \begin{subfigure}{\linewidth}
        \centering
        \includegraphics[width=0.8\textwidth]{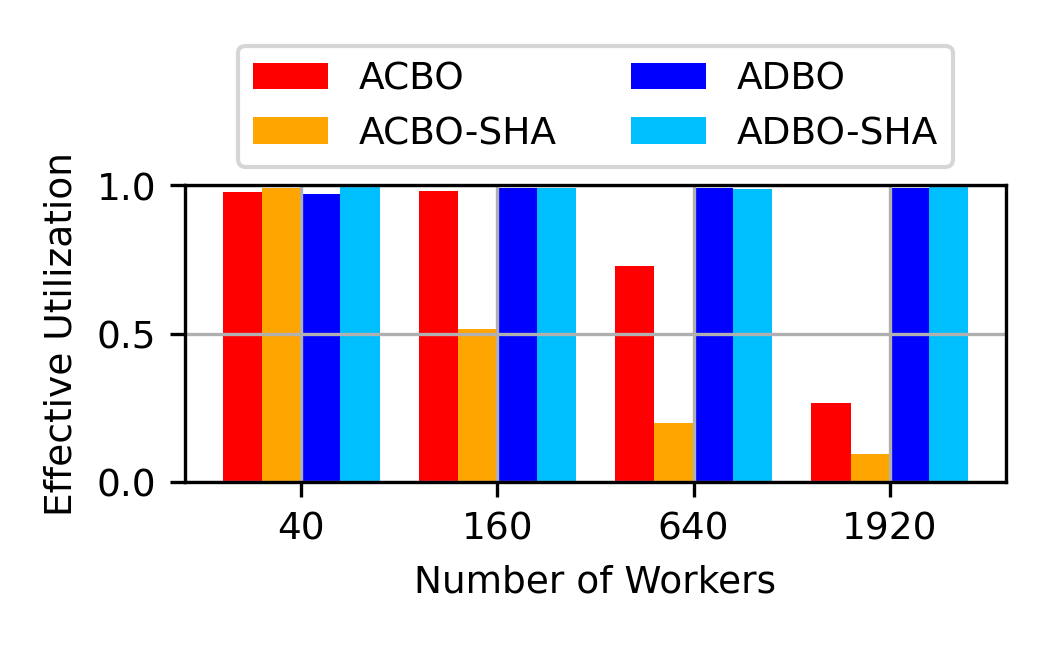}
        \vspace{-0.4cm}
        \caption{Effective utilization of workers.}
        \label{fig:scaling-utilization}
    \end{subfigure}
    \begin{subfigure}{\linewidth}
        \centering
        \includegraphics[width=0.8\textwidth]{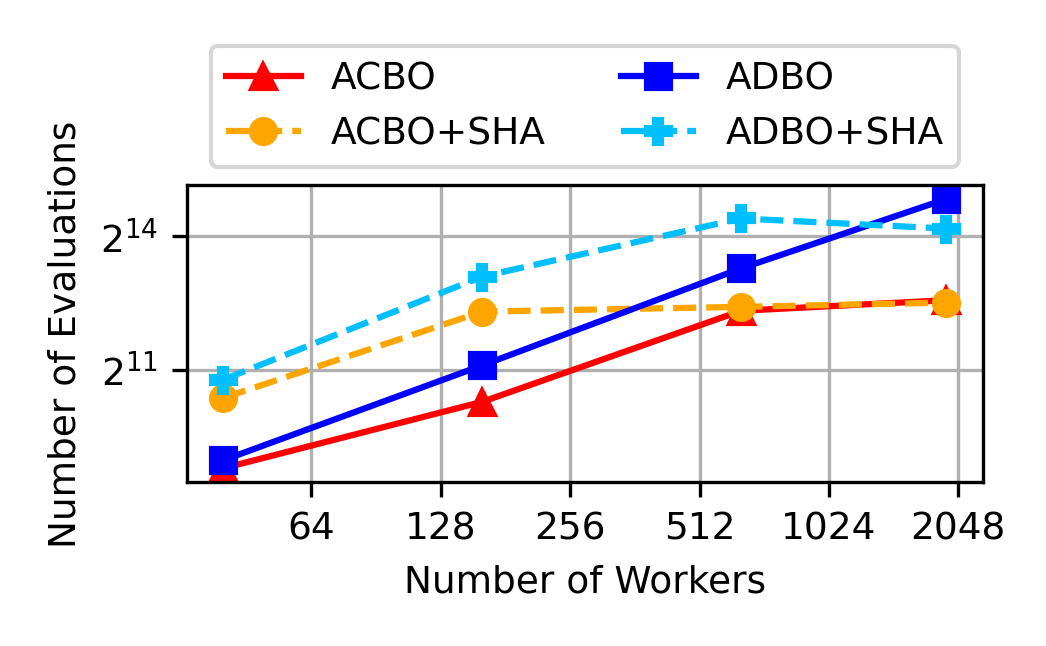}
        \vspace{-0.4cm}
        \caption{Number of evaluated hyperparameter configurations.}
        \label{fig:scaling-num-evaluations}
    \end{subfigure}
    \caption{Comparison of (\ref{fig:scaling-utilization}) effective utilization and (\ref{fig:scaling-num-evaluations}) number of completed hyperparameter evaluations of centralized/decentralized BO without/with Successive Halving (SHA) for an increasing number of parallel workers (1 GPU per worker). Overall, {\color[HTML]{3166FF} the decentralized approach is better with higher utilization and the number of evaluated configurations}. {Conversely, \color[HTML]{CB0000} the utilization and number of evaluations quickly drop for the centralized approach}. }
    \label{fig:computational-performance}
    \vspace{-0.4cm}
\end{figure}

We study the effectiveness of the centralized and decentralized architectures when the number of workers is increased. The general idea of looking at resource utilization  is to detect overheads in parallel algorithms which when excessive can be counterproductive and worsen the results even though resources were increased. For this, we compute the effective utilization which we defined as the ratio between the cumulative time spent in training neural networks and the allocated total time ($= \text{Number of Workers} \times \text{Execution Time}$). The resource utilization w.r.t. the number of parallel workers is  presented in Figure~\ref{fig:scaling-utilization}. The centralized architecture has a utilization similar ($> 90\%$) to the decentralized at a small scale (40 workers). However, when the number of workers increases utilization drops quickly and finishes below 25\% meaning that more than 75\% of allocated resources are not being used. In addition, using an early discarding strategy makes the computational efficiency even worse as it can be observed that from 160 workers Cent. SHA (orange) has about half the utilization of Centralized (red). This behavior is normal as early discarding shortens the duration of evaluations by stopping early non-promising networks which increases the number of queries received by the manager, resulting in more frequent overhead. 

But, looking at the utilization is not sufficient as it could be kept artificially high just by submitting hyperparameter configurations which take longer to be completed and therefore bypass the problem of querying frequently the BO agent. Therefore, we propose to also look at the number of completed evaluations, which is shown in Figure~\ref{fig:scaling-num-evaluations}. The number of completed evaluations is significantly higher for decentralized executions than centralized ones. For 1,920 workers, the decentralized completed 29,222 evaluations while the centralized completed only 6,055 evaluations. Similarly Dist. SHA completed 18,431 while Cent. SHA completed only 5,832 evaluations. The number of evaluations increases linearly for the decentralized, unlike the Centralized and Cent. SHA which plateau. The behavior of Dist. SHA with 1,920 workers can be explained through the BO agent which is suggesting in this case hyperparameter configurations that are longer to train. By looking at the number of evaluations we showed that under frequent queries the search can maintain high effective utilization.

\begin{figure} 
    \centering
    \includegraphics[width=0.8\columnwidth]{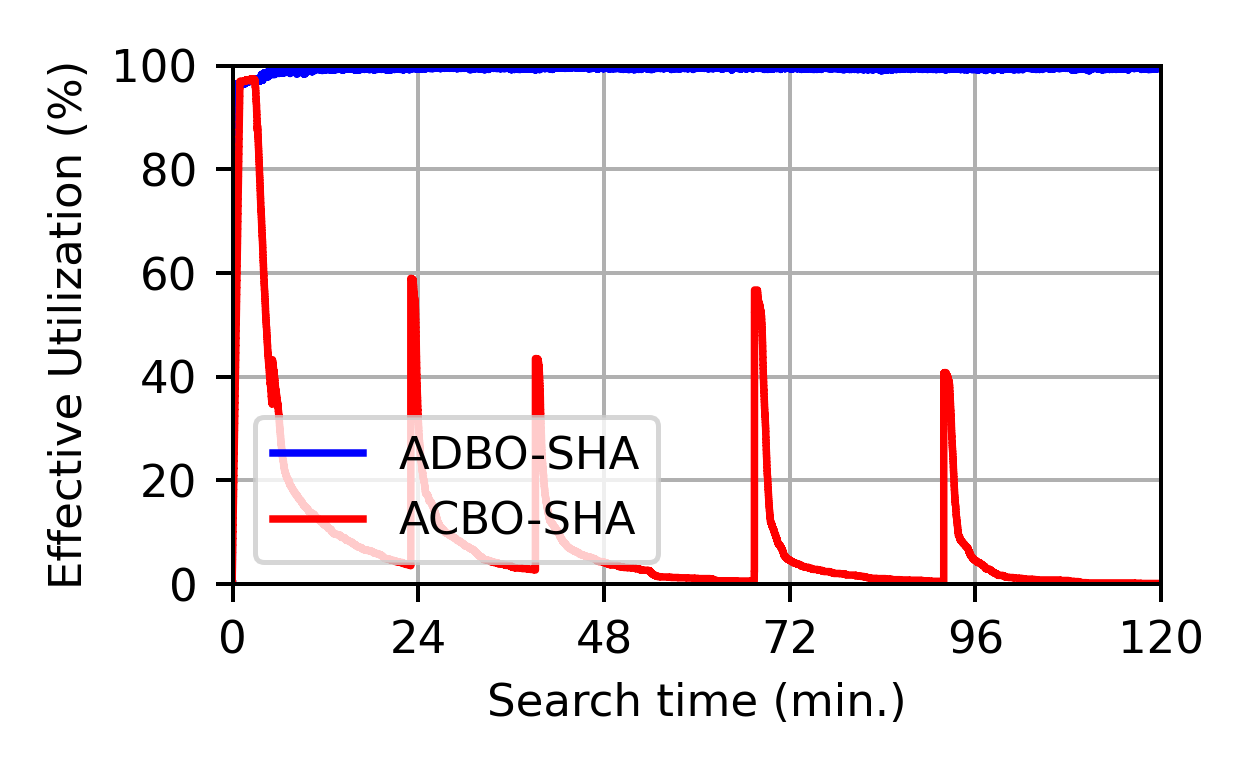}
    \vspace{-0.4cm}
    \caption{Comparing the evolution of worker utilization for the centralized and decentralized BO with a total of 1,920 parallel workers. Overall, {\color[HTML]{3166FF} the decentralized approach has high utilization} while {\color[HTML]{CB0000} the centralized approach drops early on without ever recovering}.}
    \vspace{-0.4cm}
    \label{fig:cbo-vs-dbo-utilization}
\end{figure}

\begin{figure*}[!ht]
    \centering
    \begin{subfigure}{\columnwidth}
        \centering
        \includegraphics[width=0.8\textwidth]{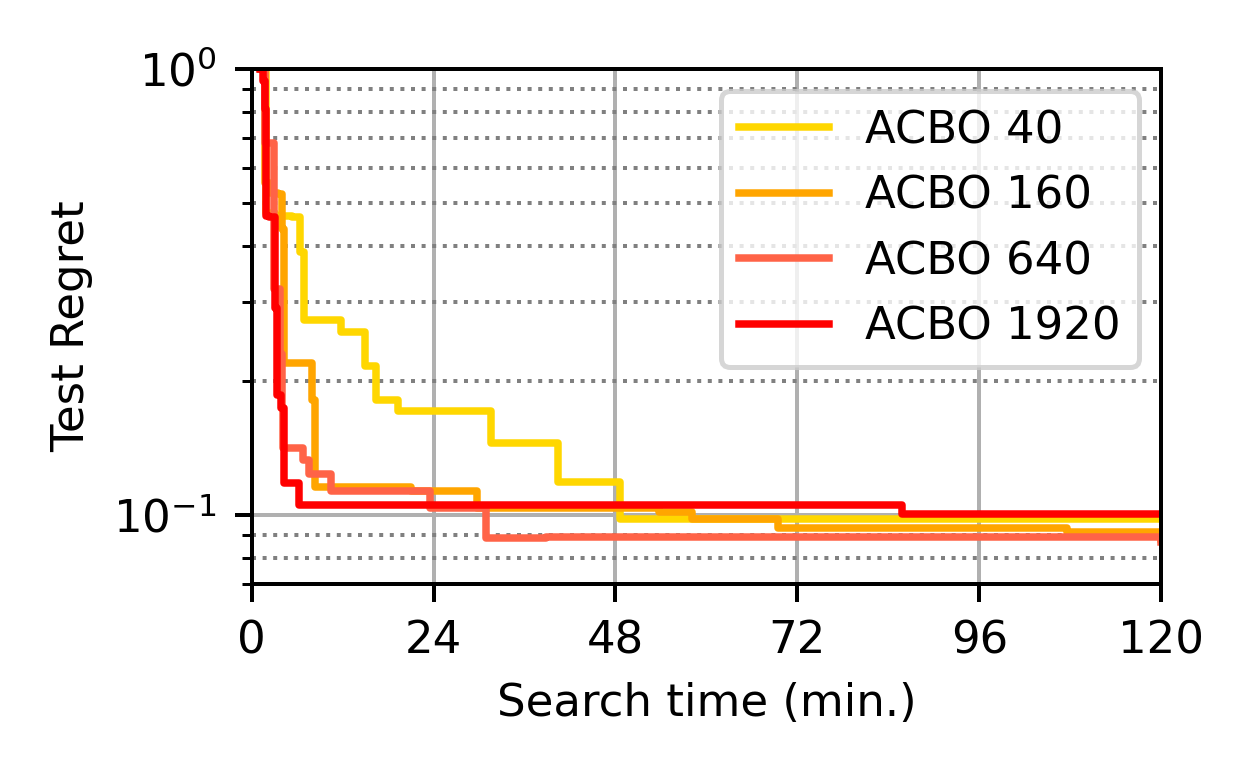}
        \vspace{-0.4cm}
        \caption{Centralized}
        \label{fig:cbo-objective}
    \end{subfigure}
    \begin{subfigure}{\columnwidth}
        \centering
        \includegraphics[width=0.8\textwidth]{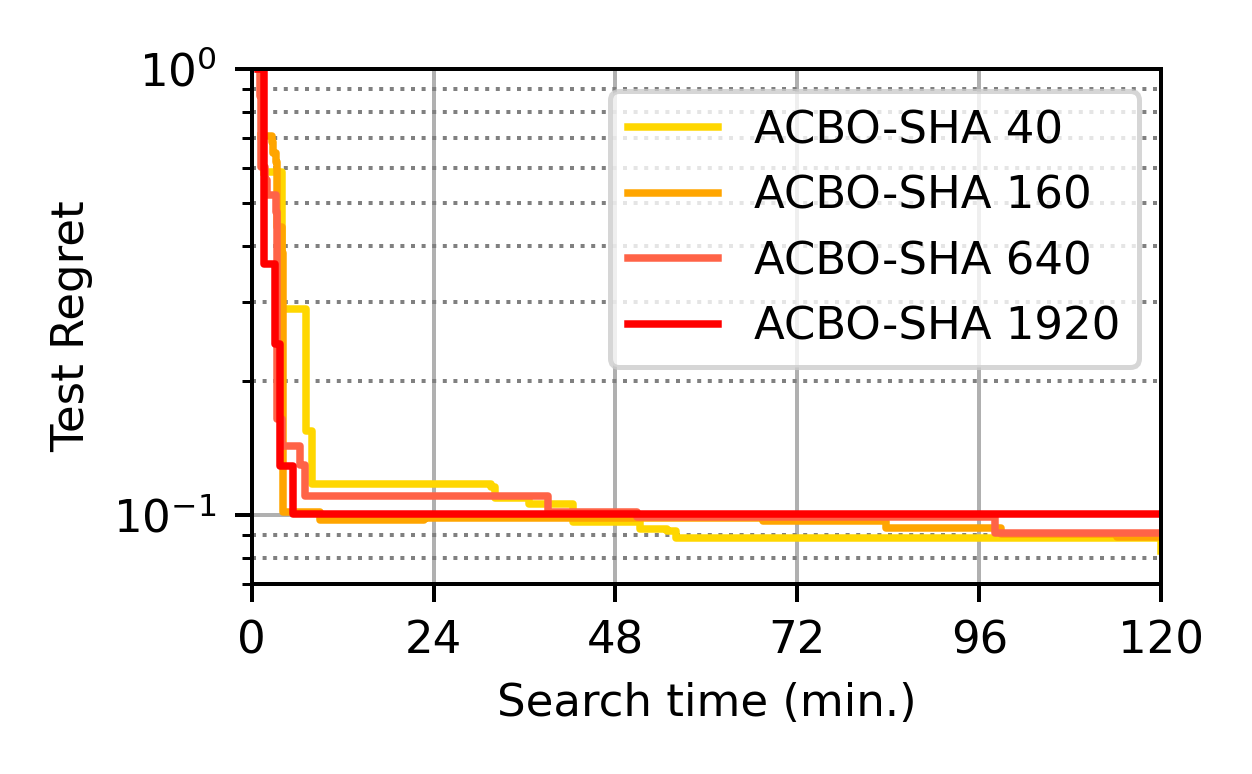}
        \vspace{-0.4cm}
        \caption{Centralized SHA}
        \label{fig:cbosha-objective}
    \end{subfigure}
    \begin{subfigure}{\columnwidth}
        \centering
        \includegraphics[width=0.8\textwidth]{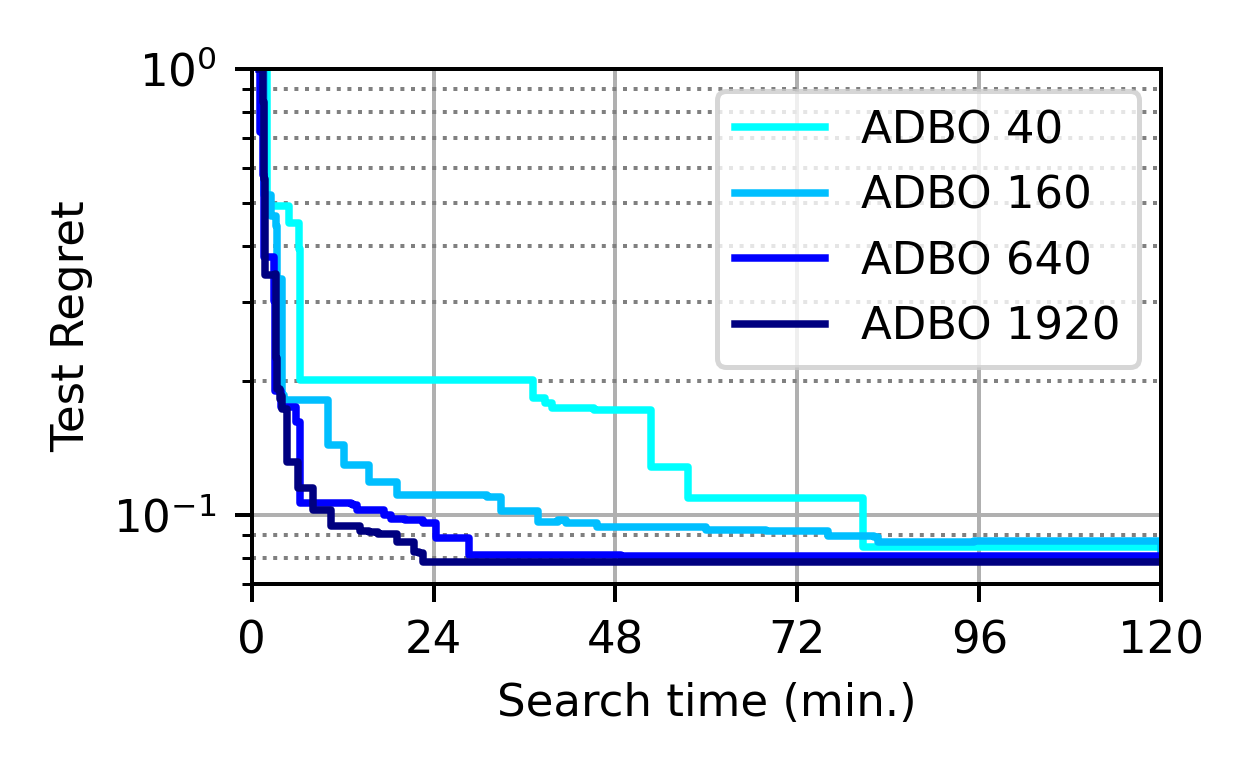}
        \vspace{-0.4cm}
        \caption{Decentralized}
        \label{fig:dbo-objective}
    \end{subfigure}
    \begin{subfigure}{\columnwidth}
        \centering
        \includegraphics[width=0.8\textwidth]{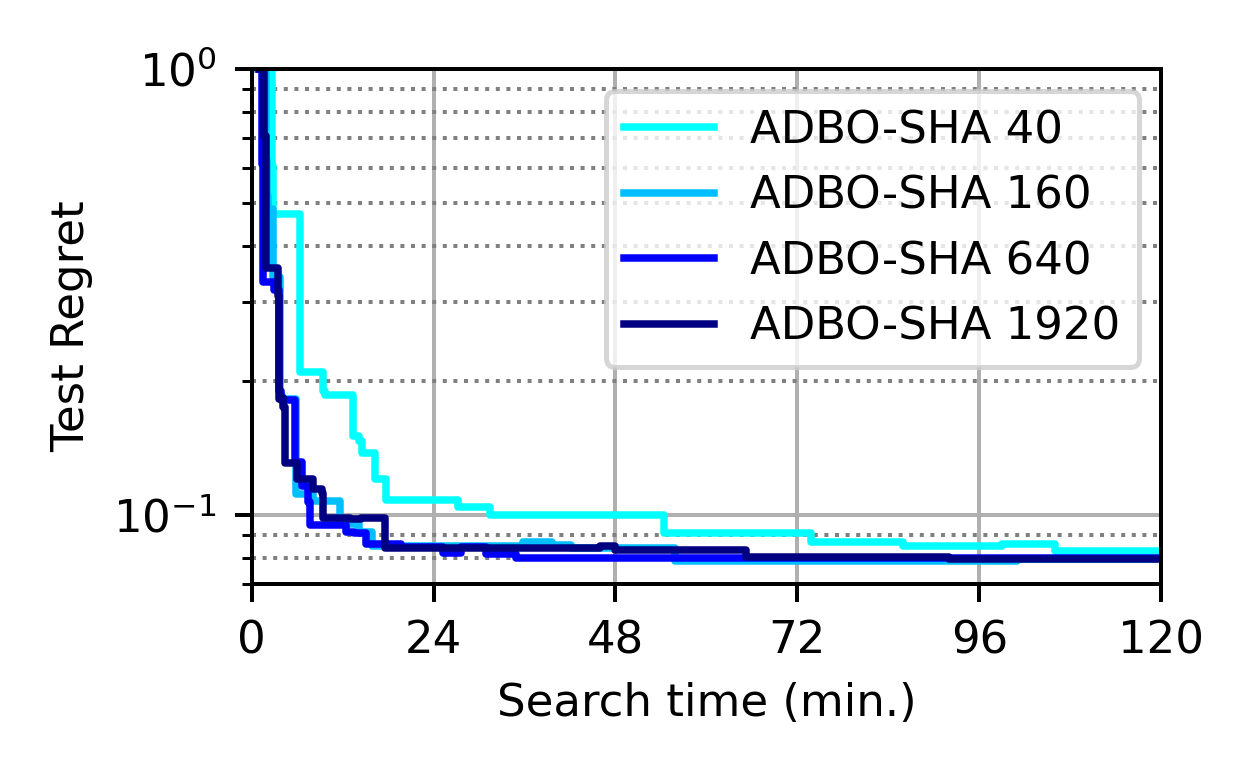}
        \vspace{-0.4cm}
        \caption{Decentralized SHA}
        \label{fig:dbosha-objective}
    \end{subfigure}
    \caption{Comparing the ``search trajectory'' (i.e., the evolution of the test regret w.r.t. the execution time) for the centralized (top, \ref{fig:cbo-objective} and \ref{fig:cbosha-objective}) and decentralized (bottom, \ref{fig:dbo-objective} and \ref{fig:dbosha-objective}) implementations with and without successive halving (SHA). Overall, {\color[HTML]{3166FF} the decentralized scales consistently where each search trajectory with a larger number of workers dominates or performs similarly to smaller scales}. A similar performance in regret can be explained by the benchmark reaching saturation (i.e., no better objective can be found). On the contrary, {\color[HTML]{CB0000} the centralized approach becomes inefficient at large scales and performs worse than at smaller scales}.}
    \label{fig:objective}
    \vspace{-0.4cm}
\end{figure*}

Finally, to explain the drop in utilization of centralized executions we take a closer look at the profile of the utilization during execution. The profiles are presented in Figure~\ref{fig:cbo-vs-dbo-utilization} and Figure~\ref{fig:cbosha-vs-dbosha-utilization} respectively for the black-box and gray box settings. It can be observed that the Centralized starts to lag after only 10 minutes of execution (when the first results are received) and display some oscillations later on which are the results of completed batches of received queries. However, these queries come in too quickly to be processed and the manager is overloaded which results in congestion and a drop in utilization that can never be recovered. The same type of profile can be observed in the Cent. SHA case as well as in previous literature~\cite{balaprakash_deephyper_2018}.

\subsection{Search Quality}
\label{sec:exp-search-quality}

After analyzing the computational performance of ADBO compared to ACBO we now evaluate the gain in ``search quality'' when the number of workers is increased. Increasing the quantity of computational resource should speed up the HPO process as well as improve the final results (i.e., the returned solution has better predictive capabilities).

The first thing we analyze is the ``search trajectory'' which corresponds to the evolution of the test regret w.r.t. the execution time such as presented in Figure~\ref{fig:objective}. Intuitively, the search trajectory can help judge the quality of an HPO algorithm through at least two aspects: first the solution after convergence, and second the time it takes to reach this solution. We start by providing a qualitative analysis of these figures which is later supported by quantitative metrics. It is clear from the centralized setting in Figure~\ref{fig:cbo-objective} and \ref{fig:cbosha-objective} that increasing the number of workers has a negative impact on the quality of the search. Indeed, when using 1,920 workers even though the early iterations are efficient, which confirms the ``sample efficiency'' of the constant-liar strategy, the trajectory reaches a plateau early on which is surpassed by executions with fewer workers. However, in the case of the decentralized setting in Figure~\ref{fig:dbo-objective} and Figure~\ref{fig:dbosha-objective} the trajectories corresponding to a larger number of workers are dominating trajectories with fewer workers. It means that it is faster to get to the same solution and the final solution is better.

\begin{figure*} 
    \centering
    \begin{subfigure}{\columnwidth}
        \centering
        \includegraphics[width=0.8\textwidth]{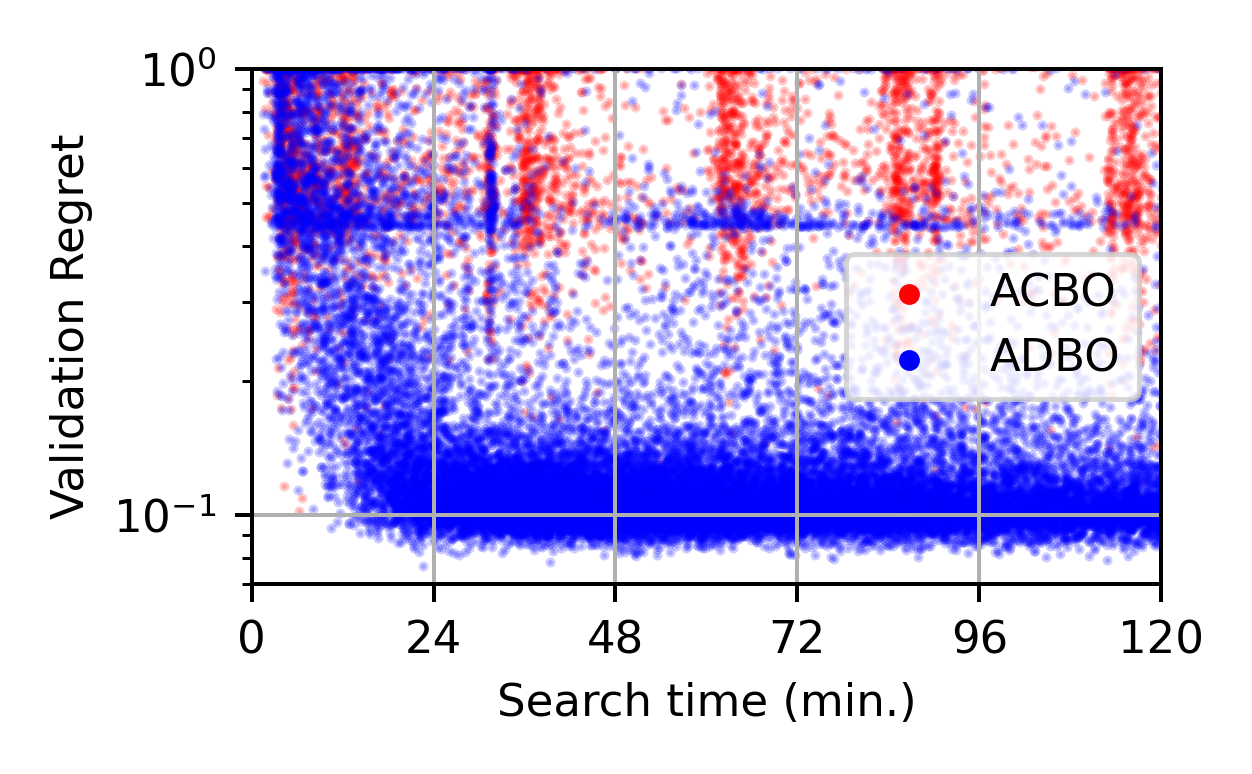}
        \vspace{-0.4cm}
        \caption{Black-Box Bayesian Optimization.}
        \label{fig:cbo-vs-dbo-scatter}
    \end{subfigure}
    \begin{subfigure}{\columnwidth}
        \centering
        \includegraphics[width=0.8\textwidth]{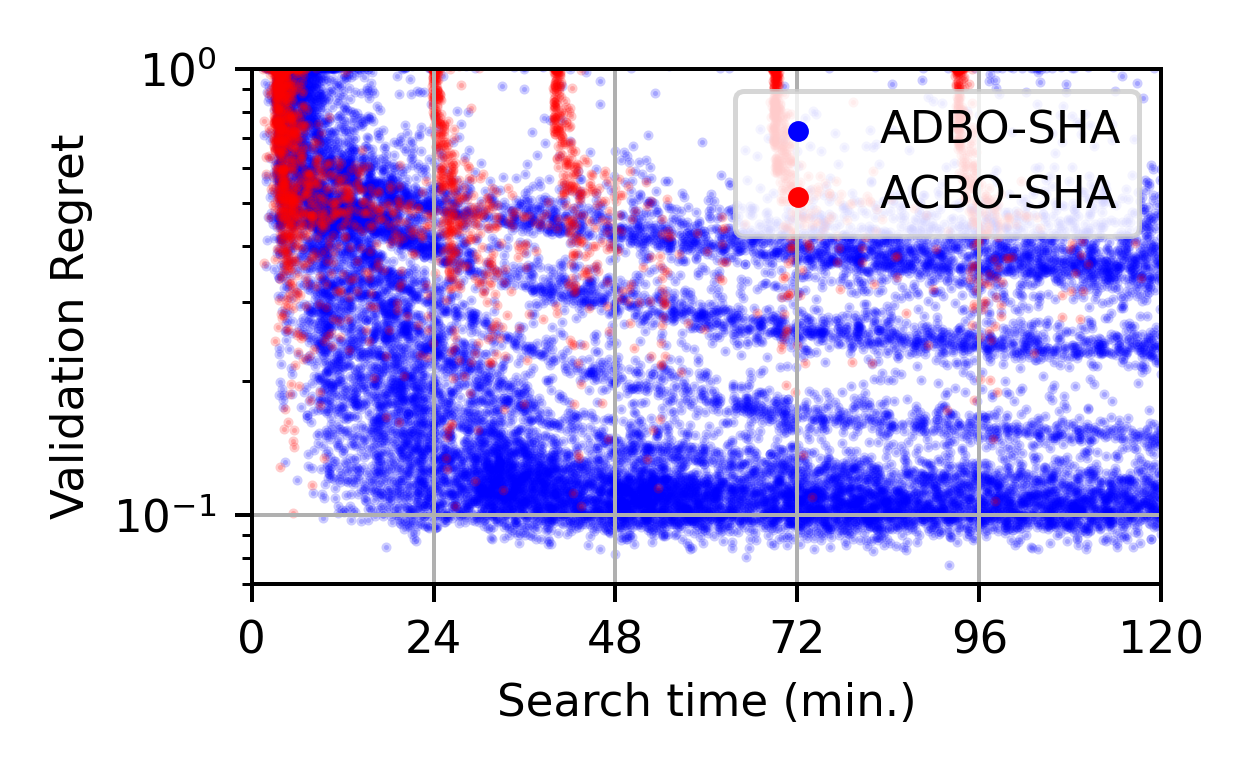}
        \vspace{-0.4cm}
        \caption{Gray-Box Bayesian Optimization.}
        \label{fig:cbosha-vs-dbosha-scatter}
    \end{subfigure}
    \caption{Comparing the observations of centralized (red) and decentralized (blue) approaches for both (a) black-box and (b) gray-box (with Successive Halving -- SHA) settings at a scale of 1,920 parallel workers.}
    \label{fig:scatter-plots-large-scales}
    \vspace{-0.4cm}
\end{figure*}

\begin{table*} 
\vspace{0.5cm}
\centering
\resizebox{0.9\linewidth}{!}{%
\begin{tabular}{c|cc|cc|cc|cc|}
\cline{2-9}
                                                         & \multicolumn{2}{c|}{\textbf{40 Workers}}                                      & \multicolumn{2}{c|}{\textbf{160 Workers}}                                     & \multicolumn{2}{c|}{\textbf{640 Workers}}                                    & \multicolumn{2}{c|}{\textbf{1920 Workers}}                                    \\ \cline{1-1}
\multicolumn{1}{|c|}{\textbf{Method}}                    & AURC                                  & Min. Regret                            & AURC                                  & Min. Regret                            & AURC                                  & Min. Regret                           & AURC                                  & Min. Regret                            \\ \hline
\multicolumn{1}{|c|}{{\color[HTML]{000000} Random}}      & {\color[HTML]{000000} 0.234}          & {\color[HTML]{000000} 0.185}          & {\color[HTML]{000000} 0.156}          & {\color[HTML]{000000} 0.116}          & {\color[HTML]{000000} 0.128}          & {\color[HTML]{000000} 0.104}         & 0.12                                    & 0.096                                    \\
\multicolumn{1}{|c|}{{\color[HTML]{CB0000} ACBO}} & {\color[HTML]{CB0000} 0.154}          & {\color[HTML]{CB0000} 0.098}          & {\color[HTML]{CB0000} 0.126}          & {\color[HTML]{CB0000} 0.092}          & {\color[HTML]{CB0000} 0.116}          & {\color[HTML]{CB0000} 0.089}         & {\color[HTML]{CB0000} 0.122}          & {\color[HTML]{CB0000} 0.100}          \\
\multicolumn{1}{|c|}{{\color[HTML]{3166FF} ADBO}} & {\color[HTML]{3166FF} 0.162}          & {\color[HTML]{3166FF} 0.085}          & {\color[HTML]{3166FF} 0.120}          & {\color[HTML]{3166FF} 0.087}          & {\color[HTML]{3166FF} 0.102}          & {\color[HTML]{3166FF} 0.081}         & {\color[HTML]{3166FF} \textbf{0.099}} & {\color[HTML]{3166FF} \textbf{0.079}} \\
\multicolumn{1}{|c|}{{\color[HTML]{CB0000} ACBO-SHA}}   & {\color[HTML]{CB0000} \textbf{0.125}} & {\color[HTML]{CB0000} 0.089}          & {\color[HTML]{CB0000} 0.118}          & {\color[HTML]{CB0000} 0.089}          & {\color[HTML]{CB0000} 0.119}          & {\color[HTML]{CB0000} 0.091}         & {\color[HTML]{CB0000} 0.117}          & {\color[HTML]{CB0000} 0.1}            \\
\multicolumn{1}{|c|}{{\color[HTML]{3166FF} ADBO-SHA}}   & {\color[HTML]{3166FF} 0.132}          & {\color[HTML]{3166FF} \textbf{0.083}} & {\color[HTML]{3166FF} \textbf{0.105}} & {\color[HTML]{3166FF} \textbf{0.080}} & {\color[HTML]{3166FF} \textbf{0.099}} & {\color[HTML]{3166FF} \textbf{0.08}} & {\color[HTML]{3166FF} 0.103}          & {\color[HTML]{3166FF} 0.08}           \\ \hline
\end{tabular}%
}
\caption{Summarizing the results across all the experiments for the Combo Benchmark. The {\color[HTML]{CB0000}centralized is colored in red} and the {\color[HTML]{3166FF}decentralized is colored in blue}. The \textit{``AURC''} represents the  Area Under the (test) Regret Curve w.r.t. the time. The \textit{``Min. Regret''} is the value of the test regret corresponding to the best validation objective observed by the search. Smaller values of AURC and Min. Regret are better. The best score across methods are in bold font. Overall, {\bf \color[HTML]{CB0000} at the smallest scale the centralized is the most efficient (smallest AURC)} which confirms the sample efficiency of the constant-liar strategy. However, {\bf \color[HTML]{3166FF}when increasing the scale the decentralized is better} with faster convergence and smaller regret.}
\label{tab:all-results}
\vspace{-0.4cm}
\end{table*}

Now, the search trajectory is not all we want to observe because it does not display the ``strength'' of convergence but only the quality of iterative improvements. Therefore, the online observations of the execution at the largest scales are presented through scatter plots in Figure~\ref{fig:cbo-vs-dbo-scatter} and \ref{fig:cbosha-vs-dbosha-scatter}. Through these figures, it is clear that the decentralized variants converged to an area of the search space consistently suggesting better hyperparameter configurations (the baseline has a performance of 0.125 in test regret). On the contrary, the centralized variants seem to rarely suggest good configurations. The capability of a search to discover different hyperparameter configurations and therefore neural networks architectures can be used to assess the uncertainty in model-choice (aka epistemic uncertainty) and reduce the variance of estimation through ``ensembling'' such as proposed in~\cite{wenzel2020hyperparameter,egele2022autodeuq}. The behavior of SHA can also be observed through the stratification of the y-axis in Figure~\ref{fig:cbosha-vs-dbosha-scatter} (i.e., early termination of low-performing configurations).

Finally, from a quantitative point of view we summarize the AURC and minimum test regret for all the methods in Table~\ref{tab:all-results} over the different scales of parallel workers. The performance of random search (without early discarding) is presented to complement our previous sanity checks. Also, random search scales without issues as communication is not required and is a good competitor when increasing the number of workers. At a small scale (40 workers) the centralized has a smaller AURC with a regret very close to the decentralized. This confirms that at a small workers' scale, the centralized approach is efficient and legitimate. Then, when increasing the workers (from 160 to 1,920) the decentralized consistently outperform the centralized settings. Also interesting, when we reach the largest scale of workers we enter in a regime where ``compute'' is not a bottleneck for HPO anymore and we can observe that ADBO (maximum fidelity) is outperforming ADBO+SHA (early discarding).

\section{Conclusion}

The issue we address is the task of optimizing resource utilization for large scale HPO. We consider this problem to be of importance based on the last generation of released HPC systems that are equipped with at least 1,000 accelerator chips (e.g., GPUs) and the popularity of ML in science. In this study, we focus on a well-adopted BO procedure, based on a Random-Forest surrogate model~\cite{hutter2011sequential}, which already demonstrated its efficiency  to solve the HPO problem both sequentially and at small scales of parallelism. Then, we focus on the asynchronous querying of the BO agent as the training time of neural networks can vary depending on their hyperparameter configurations (e.g., learning rate, batch size, number of weights) and it was already shown to improve worker utilization~\cite{li2020system}. We study the role of centralized and decentralized architectures of parallel BO. Indeed, centralized approaches are known to be more sample-efficient.
However, the agent quickly becomes overloaded when the number of workers increases 
Therefore we developed a decentralized BO procedure with a custom acquisition function heuristic to maintain efficiency. We demonstrated at large scales (up to 1,920 workers), both in the case of black-box and gray-box optimization, that this approach improves significantly the utilization of computational resources compared to the centralized setting. We presented the profiles of worker utilization to explain our results. Finally, we summarized our results over different scales of parallel workers with scalar metrics (the ``Area Under the Regret Curve'' and the ``Minimum Regret'') to demonstrate the advantage both in the speed of convergence and quality of the solution when using the decentralized BO.

During our investigations we considered for comparison the most popular available frameworks for distributed HPO such as RayTune~\cite{liaw2018tune} and Optuna~\cite{optuna_software}. Both had issues when scaling to the full Polaris system. For RayTune, the issue was to initialize the Ray cluster which resulted in a significant overhead and with many workers failing to connect to the main server. For Optuna configured with PostgreSQL and TPE sampler, when running properly at smaller scales (up to 640 workers) the best objective was similar to our proposed ADBO but resource utilization was always significantly lower and impacted by the scale. In this case, the limiting factor triggering failures was the number of parallel open connections to the database (larger than the number of workers).

One limitation of this study is the metric of worker utilization, measuring the time spent computing the black-box function which does not accurately reflect all the bottlenecks of the BO procedure when increasing workers such as the overheads associated with parallel training of neural networks. These overheads, such as parallel I/O (e.g., reading data and checkpoint weights), can create latency that is not captured in the worker utilization metric. Another important limitation of the paper is that the speed-up achieved by scaling workers in HPO is problem-dependent and may vary based on factors such as the dataset, hyperparameter search space, and ML pipeline (e.g., maximum number of training epochs). Furthermore, the lack of repeated experiments is also to be noted with the known variability of training neural networks making the results in objective quality inconclusive.
Lastly, the paper does not evaluate Gaussian processes regression due to their quick limitations w.r.t. the number of observations. Overall, this paper provides insights into the challenges associated with scaling the number of workers for HPO using BO procedures.

We are making our work available to the community as part of existing open-source software\footnote{\href{http://deephyper.readthedocs.io}{deephyper.readthedocs.io}}. In our future work, we aim to transfer the discovered benefits of this study to other applications such as simulation calibration, software, and workflow tuning. Other areas of improvement are 1) the use of accelerators for surrogate model updates, 2) low overhead optimization of the acquisition function instead of random sampling, 3) domain decomposition of the search space, 4) optimization of parallel I/O due to repeated neural network training, and 5) multi-objective optimization.

\section{Acknowledgment}
This material is based upon work supported by the U.S.\ Department of Energy 
(DOE), Office of Science, Office of Advanced Scientific Computing Research, under
Contract DE-AC02-06CH11357. This research used resources of the Argonne 
Leadership Computing Facility, which is a DOE Office of Science User Facility. 
This material is based upon work supported by ANR Chair of Artificial Intelligence HUMANIA ANR-19-CHIA-0022 and TAILOR EU Horizon 2020 grant 952215.

\bibliographystyle{IEEEtran}
\bibliography{references,zotero-romain}



\end{document}